\documentclass[runningheads]{llncs}
\usepackage{cite}
\usepackage{amsmath,amssymb,amsfonts}
\usepackage{multirow}
\usepackage[table,xcdraw]{xcolor}
\usepackage{adjustbox}
\usepackage{makecell}
\usepackage{algorithmic}
\usepackage{subcaption}
\usepackage{graphicx}
\usepackage{url}
\usepackage[colorlinks,urlcolor=blue]{hyperref}

\usepackage{comment}
\usepackage{fix-cm}
\usepackage{textcomp}
\usepackage{float}
\usepackage{svg}
\usepackage{caption}
\usepackage[T1]{fontenc}
\DeclareCaptionFont{xxviii}{\fontsize{9}{9}\selectfont}
\renewcommand{\tablename}{Table}
\renewcommand{\thetable}{\arabic{table}}
\cellcolor[HTML]{D7D6D6}
\def\BibTeX{{\rm B\kern-.05em{\sc i\kern-.025em b}\kern-.08em
    T\kern-.1667em\lower.7ex\hbox{E}\kern-.125emX}}

\begin{document}

\title{RailSafeNet: Visual Scene Understanding for Tram Safety}
\author{Ondřej Valach\inst{1}\orcidID{0009-0000-7629-0516}, \and Ivan Gruber\inst{1}\orcidID{0000-0003-2333-433X}}
\institute{Department of Cybernetics, University of West Bohemia, Pilsen, Czechia\\
\email{valacho@kky.zcu.cz, grubiv@ntis.zcu.cz}}

\maketitle

\begin{abstract}
Tram-human interaction safety is an important challenge,
given that trams frequently operate in densely populated
areas, where collisions can range from minor injuries to fatal
outcomes. This paper addresses the issue from the perspective
of designing a solution leveraging digital image processing,
deep learning, and artificial intelligence to improve the safety
of pedestrians, drivers, cyclists, pets, and tram passengers. 
We present RailSafeNet, a real‑time framework that fuses semantic segmentation, object detection and a rule‑based Distance Assessor to highlight track intrusions. Using only monocular video, the system identifies rails, localises nearby objects and classifies their risk by comparing projected distances with the standard 1435mm rail gauge. Experiments on the diverse RailSem19 dataset show that a class‑filtered SegFormer B3 model achieves 65\% intersection-over-union (IoU), while a fine‑tuned YOLOv8 attains 75.6\% mean average precision (mAP) calculated at an intersection over union (IoU) threshold of 0.50. RailSafeNet therefore delivers accurate, annotation‑light scene understanding that can warn drivers before dangerous situations escalate. Code available at \url{https://github.com/oValach/RailSafeNet}.


\end{abstract}

\section{Introduction}\label{intro}
Urban tram networks remain prone to severe conflicts with other road users. A~recent study of 7,535 incidents across Germany, Austria, Switzerland and Sweden recorded 8,802 injuries; although only 3\% proved fatal, almost one-quarter were classified as serious~\cite{lackner2022tram}. Most crashes involve pedestrians or cyclists who share the track corridor, and the typical impact pattern—head, chest and lower-limb trauma—points to the limited protection these users enjoy. Reducing such harm therefore demands not only better infrastructure and traffic rules~\cite{guerrieri2018tramways} but also on-board assistance able to warn drivers before a collision happens.

\begin{figure}[!t]
    \centering
    \includegraphics[width=\textwidth]{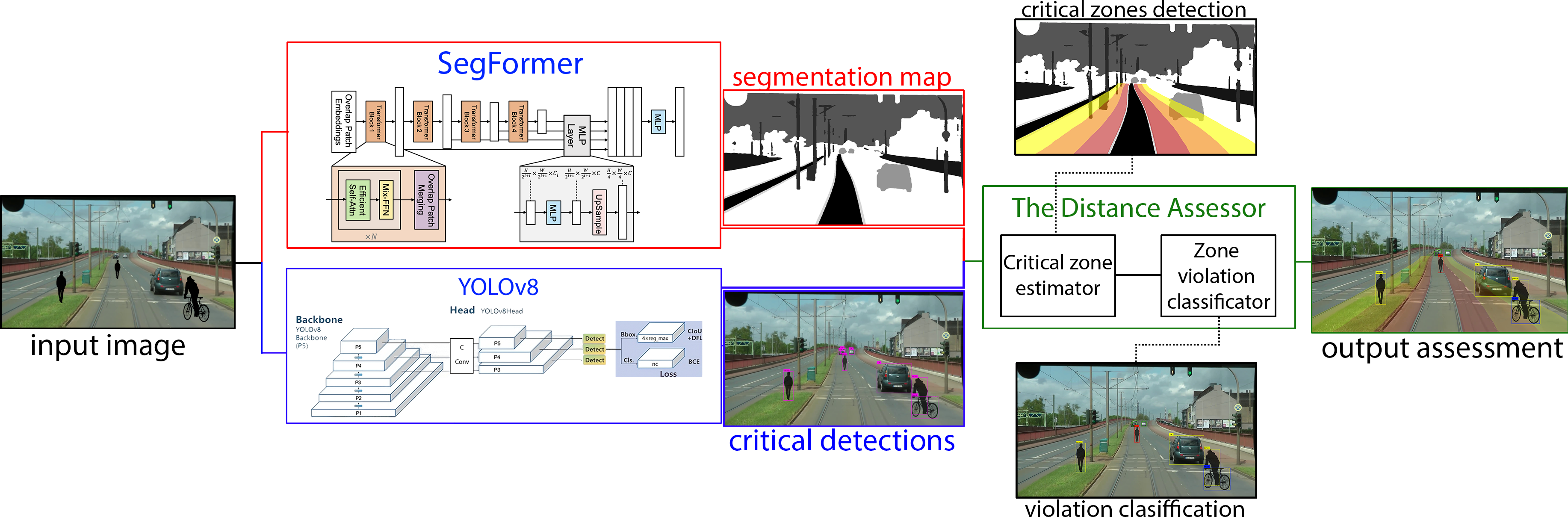}
    \captionsetup{width=\textwidth}
    \caption{The system pipeline visualization, starting with the system input data sample on the left, continuing to the parallel segmentation and detection, entering The Distance Assessor and ultimately coming out of the system as an output visualization.}
    \label{arch}
\end{figure}

Recent progress in computer vision makes that goal realistic. Real-time semantic segmentation can now delineate rail geometry, while lightweight detectors localize people, cars and other critical objects in the same frame. What is still missing is a monocular method that converts those predictions into an accurate, camera-agnostic estimate of an object’s distance from the track. We solve this problem with a pipeline (see Figure~\ref{arch}) that: (i) segments rails, (ii) detects surrounding objects, and (iii) infers distance via rule described in Section~\ref{assessor}. By fusing these steps in a rule-based \emph{Distance Assessor}, our system turns the driver’s forward view into a continuously updated “criticality map” highlighting objects that could intrude into the tram path.

The proposed approach requires no camera calibration, depth sensor or LiDAR, and it runs fast enough for in-cab deployment. Ultimately, it provides tram operators with a low-cost means to mitigate collisions with pedestrians, cyclists, vehicles, animals or debris before they become unavoidable.

The three main contributions of this paper are as follows:  (i) we introduce an approach for the task of \textbf{distance to rail estimation} in images without any knowledge of the capture, camera or setup parameters from \textbf{single image input}, (ii) we utilize a \textbf{custom segmentation processing} approach using data class filtering and mask post-processing, achieving superior results compared to the original RailSem19 dataset paper, (iii) we propose a \textbf{Distance Assessor system} which processes outputs from scene segmentation and object detection to accurately estimate distances from track and classify object criticality.

\section{Related Work}

Interactions between trams, pedestrians and vehicles still generate disproportionate numbers of serious pedestrian injuries and fatalities~\cite{Purs2024TramPedestrianCM,lackner2022tram,Jedk2023VariousAT,Lackner2022TramTP}. Collision analyses further show that tram-front design and risky pedestrian behavior amplify trauma severity~\cite{Jedk2023VariousAT,Lackner2023InvestigationsOT,Lopot2023PedestrianSI}.

Onboard perception has advanced with computer-vision progress. FCN and U-Net pioneered dense prediction~\cite{ronneberger2015u}; DeepLabv3+ added multi-scale atrous pooling~\cite{chen2017deeplab}; transformer models Mask2Former and SegFormer give lightweight real-time masks~\cite{cheng2022masked,xie2021segformer}. Foundation-level SAM further boosts cross-domain accuracy~\cite{kirillov2023segment}. Object localization evolved from successive YOLO releases to transformer-based DETR~\cite{wang2024yolov9,carion2020end}.

Monocular distance estimation exploits rail geometry, linear cues or depth networks such as HybridDepth and UniDepthV2~\cite{Lee2022VehicleDE,AlHasanat2021RetinaNetBasedAF}. These cues feed prototype warning systems that issue real-time alerts~\cite{Bidve2024PotholeDM,Kozlov2024PrerequisitesFD,Shetye2023ComputerVF,Ferrer2013ImagePF}. Our study unifies these strands in a class-filtered SegFormer, fine-tuned YOLOv8 and gauge-guided Distance Assessor specialized for tram scenes.
\section{Data}\label{data}
Before detailing the methods, we first describe the RailSem19 dataset~\cite{zendel201919} used to train and evaluate our models. This dataset was selected for its large size and high variability in rail scene-related situations. It consists of 8,500 unique images extracted from 350 hours of video captured from the ego-perspective of trains and trams across 58 different countries, thereby encompassing diverse sceneries—from wilderness areas with meadows, forests, and mountains to urban environments. Unlike other available datasets such as Rail-5k~\cite{zhang2021rail}, and RailVID~\cite{yuan2022railvid} or a synthetic approach for simulating real world data~\cite{de2023scenario}, which either lack tram data, overly segment track details, or have issues with public availability, RailSem19 provides segmented classes and combined train/tram scenarios along with annotations, including semantic segmentation dense label masks and JSON files containing rail-relevant polygons and rails as polylines. This dataset also includes bounding box annotations, these are further utilized for fine-tuning, see more in Section~\ref{detection}.

\begin{figure}[!t]
    \centering
    \includegraphics[width=0.8\textwidth]{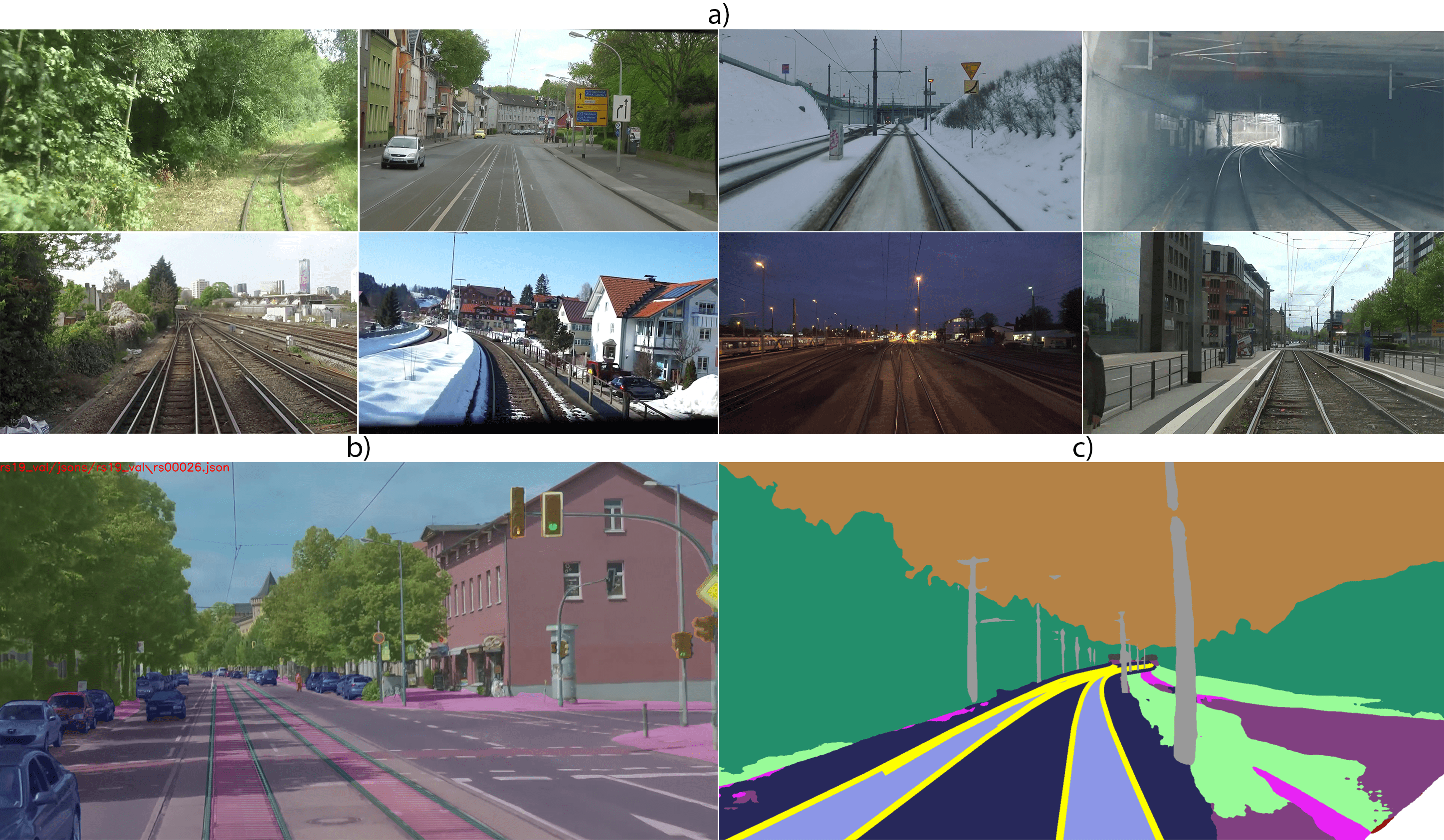}
    \caption{RailSem19~\cite{zendel201919} dataset examples, eight different data samples showing diversity in different ways (a), example segmentation mask in a photo overlay (b), example segmentation mask (c)}
    \label{railsem19_examples}
\end{figure}

The dataset features 20 annotated classes, with five predominant ones, these are: \textit{construction} (8250), \textit{pole} (8200), \textit{vegetation} (8250), \textit{sky} (8150) and \textit{rail-track} (7800), its approximate occurence frequencies are in parentheses. Dataset examples showcasing environmental variability are shown in Figure~\ref{railsem19_examples}. It is essential to points out that segmentation maps were generated using a state-of-the-art semantic segmentation solution from year 2019. According to the authors of the RailSem19 paper, most of these labels are well above 90\% of intersection over union. However, after manual inspection, many misleading annotations and objects not detected or partially detected were found.

This data were split between train, test and evaluation subsets with ratio of 80/10/10 where the distribution of each subset aligns with the distribution of the whole dataset. Due to the fact of not ideally segmented ground truth segmentation masks, hand inspection of classes was performed to exclude classes with poor ground truth accuracy. This way the number of classes was reduced from twenty to eleven maintaining important classes for this project such as tram-track, rail-track and rail-raised and filtering out classes such as vegetation, terrain or road. This approach significantly increased the quality of our training data and cleaned the segmentation masks, removing the seemingly arbitrary edges in between not ideally segmented classes.

\section{Methods}
RailSafeNet consists of three main stages: rail segmentation to identify track area, object detection for critical objects localization, and distance estimation from tram to obstacle. This section describes each stage in detail.
Both image segmentation and object detection are essential for the usability of the whole system, however, both of these tasks operate independently. That can not be said about the Distance Assessor end system which relies solely on the outputs from these two models. This overall system architecture is shown in Figure~\ref{arch}.

The~image segmentation is to~be used mainly for~the~localization of~rail related classes and~scene understanding where the~output segmentation mask should include information about the~rail location, shape, direction etc. The~task of~the~object detection executed in~parallel with the~segmentation, is to detect and~provide accurate information about location and~number of~any objects that appeared to~be around the~railway. This is the~fundamental information to~work with in~the~following rule-based Distance Assessor system which should estimate the~distance of~detected objects from the~track. By combination with the~information from the~detection system, there is enough information for the distance approximation.

\subsection{Image Segmentation}\label{segmentation}
The image segmentation task provides information about the rail position in the image, based on that, the following Distance Assessor estimates zones around rails in which the tram passage could be critical to the entity present.

For this task, the chosen model is a SegFormer with tested versions B2 and B3. This model provides great trade-off between accuracy and speed, with an ability to provide real-time inference~\cite{chen2014semantic, xie2021segformer}. This high inference speed is essential for our use case. For training these SegFormer models, a class filtration and mask postprocessing (more about that in Section~\ref{eval}) was applied, 

For better model generalization, extensive image augmentations using the Albumentations~\cite{buslaev2020albumentations} library were applied with a respect to the environment structure. We evaluated three augmentation tiers: Tier 1 used the raw images with no modifications; Tier 2 introduced moderate transformations, combining geometric shifts, scaling, rotations, flips, color adjustments (brightness, contrast, jitter), blur, and coarse dropout; and Tier 3 applied the full Tier 2 with simulated weather effects (rain, fog, snow, sun flare) and extra noise effects as ISO and Gaussian noise. Augmentation Level 1 retains the original image with a probability of 12.3\%. When using Augmentation Level 2, this probability further decreases to 6.1\%.

\subsection{Object Detection}
Due to the lack of data annotations for instance segmentation in rail scene understanding, the model’s performance suffers because it is trained on multiple classes with imbalanced frequencies. Classes, such as “human” and “car”, are underrepresented compared to “rail-track” resulting in lower accuracy for critical objects. Such reduced performance was be observed, for example, by merging of heterogeneous regions, regions being incomplete or completely missing. This led to the decision of employing a separate object detection algorithm, which offers faster and more accurate object localization. A YOLOv8 model was chosen for object detection mostly because of its competitive trade-off between speed and performance~\cite{yolov8_2022}. Further, it was finetuned on RailSem19 filtered classes bounding box annotations, see more in Section~\ref{results}.

\subsection{Distance Assessor}\label{assessor}
\begin{figure}[!t]
    \centering
    \includegraphics[scale=0.13]{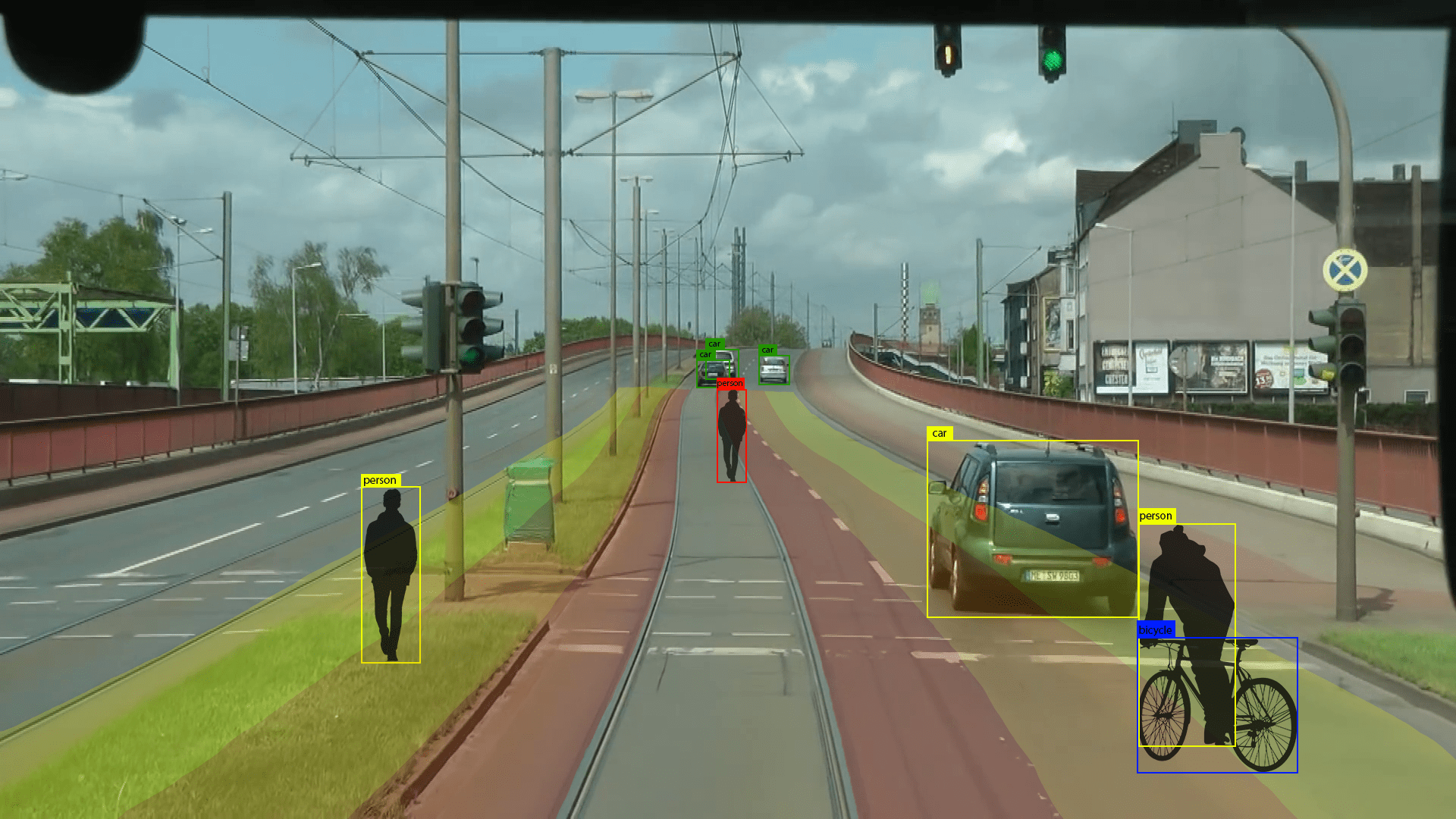}
    \caption{Visualization of the expected output from the system: each level of borders marked with different color represents different areas of the so-called ”distance criticality”. From yellow being the least critical to orange marking the moderately critical zone to red, which is the most critical zone, also including the whole rail tracks.}
    \label{vizual_res}
\end{figure}

After the segmentation mask from the SegFormer block and bounding-box predictions from the YOLOv8 block are obtained, the next step is to extract the needed information from both of these outputs. That is done in the following The Distance Assessor block which processes the segmentation mask and bounding-boxes individually. The segmentation mask is utilized to obtain estimations of the distance to rail, that happens in the Critical zones estimator sub-block. That is followed by the processing of output bounding-boxes, these are used for extraction of object locations and labeling, which happens in the sub-block Zone violation classificator. This block also evaluates the object to rail criticality based on its distance from rail utilizing the output from the Critical zone estimator. The background of The Distance Assessor block is going to be described in detail in the following paragraphs. The Distance Assessor block output should look like in Figure~\ref{vizual_res}. 

A subsystem denoted as \textbf{Critical zones estimator} leverages a key aspect of the system, that is the use of the standardized rail gauge, which measures 1435~mm across much of Europe~\cite{ciaworldfactbook2024railways}. That is a reliable reference for distance estimation. By detecting the rail in each image, the algorithm determines the number of pixels in width (\(d_{px}^{in}\)) that represent this known dimension (\(d_{real}^{in}\)). This comparison allows for the computation of a conversion factor (\(p_d\)) for each depth level affected by perspective distortion. Using this factor, the system converts distance estimations to pixel measurements following the formula

\begin{equation}\label{pixellengthindex}
D_{depth}=p_d\ d_{real}^{out};\quad p_{d}= {\frac{d_{px}^{in}}{d_{real}^{in}}},
\end{equation}

where \(d_{real}^{out}\) is the real-world distance that we need to estimate the number of pixels for. This approach enables precise distance estimation by using the rail gauge as a reference point. After the distance derivation for the given depth from each rail class mask edge point, the border between the estimated critical zone edge points is an interpolation. That way, the more depth levels are included, the more precise the final estimation will be. This way the final trio of critical zones is derived in specific distances as discussed further in Section~\ref{eval_assessor}.

Following is the classification of the objects close to the tram. The \textbf{Zone Violation Classifier} evaluates detected objects by checking whether specific border points of their bounding boxes intrude into predefined critical zones surrounding the railway. The system systematically analyzes 12 equidistant points along each bounding box - covering the corners and sides - and classifies objects based on the presence of any intrusion. Detected objects are divided into two categories: movable objects (such as persons, cars, and bicycles), which are prioritized due to their immediate life risk, and stationary objects (like backpacks and umbrellas), which are considered lower-level hazards. A color coding further helps to easily differentiate the level of criticality among the detected intrusions, using green, yellow, orange and red color in this given order starting from no risk to the highest.

\subsection{Evaluation}\label{eval}
Intersection over Union and Mean Average Precision are used for the segmentation and detection evaluation. Segmentation evaluation takes into account every class in the mask, including small meaningless patches resulting from imprecise training data (even though the class filtration). Classes representing amorphous objects like grass or terrain lack clearly defined boundaries, making precise segmentation challenging. These artifacts can be suppressed through post-processing techniques like morphological operations or rule-based filtering.

We use one technique to post-process the predicted segmentation masks, that is a morphological operation of closing (a dilation followed by an erosion), aimed to remove small meaningless patches. Because this post-processing is never flawless, we also utilize a custom modification to the IoU and mAP metrics to ignore small patches of a specific size. This patch size L was experimentally tested to be the most effective at $12 \times 12$ pixels and was utilized in our final solutions.

\section{Experiments and Results}\label{results}
This section describes the training setup and follows with training results. All models were trained using a single NVIDIA A40 GPU setup with varying learning rate and batch size, schedulers, optimizers, augmentation levels, input image sizes and number of training epochs. All of the hyperparameters and their experimented options are listed in Tab.~\ref{hyperparams}. Different combinations were deeply explored with a Wandb Sweep tool to search the hyper-parameter space, see Section~\ref{seg_exp_eval}.
Results of individual subsystems are described in following sections.

\begin{table}[!t]
\centering
\caption{Hyper-parameter ranges and settings.}
\begin{tabular}{ll}
\hline
\textbf{Model} & SegFormerB2, SegFormerB3 \\ \hline
\textbf{Learning rate} & 1e-2 – 1e-6 \\ \hline
\textbf{Batch size} & 2 – 32 \\ \hline
\textbf{No. of epochs} & 50 – 200 \\ \hline
\textbf{Scheduler} & LinearLR, ReduceLROnPlateau~\cite{paszke2017automatic} \\ \hline
\textbf{Optimizer} & SGD, Adagrad, Adam, AdamW \\ \hline
\textbf{Augmentation levels} & 0, 1, 2 \\ \hline
\textbf{Input resolution} & \begin{tabular}[c]{@{}l@{}}224×224, 480×480, 512×512, 550×550, 700×700, 1024×1024 \\ (adapted to GPU VRAM limitations along with batch size)\end{tabular} \\ \hline
\end{tabular}
\label{hyperparams}
\end{table}

\subsection{Segmentation Experimental Evaluation}\label{seg_exp_eval}
The original RailSem19 paper used the FRRNB model~\cite{pohlen2017full} for segmentation task, which serves as our baseline. However, as introduced in Section~\ref{segmentation}, we decided to explore the SegFormer architecture instead, testing different different model sizes: SegFormer B2 and SegFormer B3. Specifically, we evaluated multiple pre-trained SegFormer variants to determine the most promising starting point. These models differed in their pre-training configurations, including input resolution and dataset. We tested SegFormer B2 trained on ADE20k (512×512 input), SegFormer B2 trained on Cityscapes (1024×1024), and SegFormer B3 trained on Cityscapes (1024×1024). Without any data augmentation and using a constant learning rate with the Adam optimizer, SegFormer B2 reached the best IoU of 0.58, while B3 achieved 0.56 — both pretrained for 50 epochs on Cityscapes with 1024×1024 inputs. Based on these results, these two models were selected for further tuning using the WanDB Sweep tool. The best resulting checkpoints were then compared to the baseline model, as shown in Table~\ref{segformer_results}.

\begin{table}[!t]
    \centering
    \caption{Comparison of results for best performing SegFormer setups on RailSem19 segmentation with FRRNB baseline model trained by RailSem19 paper authors~\cite{zendel201919}. mAP and IoU metrics are reported with IoU representing evaluation with utilizing our best postprocessing approach.}
    \begin{tabular}{ccccccccc}
    \hline
    \rowcolor[HTML]{E0E0E0}
    \multirow[t]{2}{*}{\cellcolor[HTML]{E0E0E0}\textbf{Model}} &
    \multicolumn{6}{c}{\cellcolor[HTML]{E0E0E0}\textbf{Setup}} &
    \cellcolor[HTML]{E0E0E0} & \cellcolor[HTML]{E0E0E0} \\ \cline{2-7}
    \rowcolor[HTML]{E4E1E1}
    & {B} & \multicolumn{1}{l}{\cellcolor[HTML]{E0E0E0}{classes}} &
      {epochs} & {optim.} & {sched.} & {aug.} &
  \multirow{-2}{*}{\cellcolor[HTML]{E0E0E0}\textbf{mAP}} &
  \multirow{-2}{*}{\cellcolor[HTML]{E0E0E0}\textbf{\(\text{IoU}\)}} \\ \hline
    \begin{tabular}[c]{@{}l@{}}FRRNB~\cite{pohlen2017full}\end{tabular} & - & 21 & 60 & - & - & - & - & 0.576 \\ \hline
     & 2 & 11 & 80 & Adam & LRS* & 2 & 0.629 & 0.641 \\
     & 3 & 11 & 90 & Adam & RLROP & 2 & 0.634 & 0.646 \\
     & 3 & 11 & 80 & Adam & LRS* & 2 & \textbf{0.638} & \textbf{0.650} \\
    \multirow{-4}{*}{SegFormer} & 3 & 21 & 50 & Adam & LRS* & 2 & 0.587 & 0.598 \\ \hline
    \end{tabular}
    \label{segformer_results}
\end{table}

The original semantic segmentation FRRNB model achieved IoU of 0.576, our training run on all 21 classes achieved 0.563 IoU without any postprocessing and 0.598 after applying the mask postprocessing techniques (morphological operation of closing and small patches filtration). With this we surpass the baseline model by 0.022 IoU score. With training on the filtrated 11 classes, the IoU score further moves up on average by 0.047. This improvements indicates that the transformer-based model architecture in combination with our mask postprocessing techniques and extensive image augmentations, significantly contributes to superior performance in rail-track scene segmentation tasks. The best achieved IoU score with employing all techniques and picking the best found hyperparameter combination is 0.65 with the SegFormer B3. Hyperparameter searches can be accessed from following links: \href{https://api.wandb.ai/links/ovalach/c3d1w034}{0}, \href{https://api.wandb.ai/links/ovalach/y353o663}{1}, \href{https://api.wandb.ai/links/ovalach/zg6ydtro}{2}, \href{https://api.wandb.ai/links/ovalach/ktq4id2l}{3}, \href{https://api.wandb.ai/links/ovalach/zd45lzhc}{4}, \href{https://api.wandb.ai/links/ovalach/tix4fhjm}{5}, \href{https://api.wandb.ai/links/ovalach/8wouezd3}{6}, \href{https://api.wandb.ai/links/ovalach/fhkdh537}{7}.

To gain deeper insights into models performance, class-wise $IoU$ scores were evaluated and compared against a model trained on all 21 classes. As supposed, IoU scores for retained classes decrease when the model is fine-tuned even on those remaining ten classes. This indicates that the filtering strategy was effective — it didn’t just eliminate classes with low performance that were dragging down the overall IoU, but it also led to better segmentation results for the retained classes, with an average IoU improvement of 8.8\% per class.

\subsection{Object Detection Experimental Evaluation}\label{detection}

The detection model was fine-tuned using bounding-box annotations from the RailSem19 dataset. While the dataset contains a relatively high total number of class annotations, only three classes are applicable for our project's purposes: 'person' with 234 annotations, 'car' with 172 annotations, and 'truck' with 11 annotations. The majority of the remaining annotations belong to traffic signs and traffic lights, which do not carry any use for us at all. Comparison of YoloV8 baseline and fine-tuned model performance is in Table~\ref{yolo_res}. See also the fine-tuned model detections in the overall system output visualizations in Figure~\ref{outputs_figure}.
\begin{table}[H]
    \centering
    \caption{Comparison of detection results from baseline and fine-tuned YoloV8, evaluated with $\text{mAP}_{50}$ and IoU.}
    \begin{tabular}{cllcccc}
    \hline
    \rowcolor[HTML]{E0E0E0} 
    \multicolumn{3}{c}{\cellcolor[HTML]{E0E0E0}} & \multicolumn{2}{c}{\cellcolor[HTML]{E0E0E0}\textbf{Baseline}} & \multicolumn{2}{c}{\cellcolor[HTML]{E0E0E0}\textbf{Fine-tuned}} \\ \cline{4-7} 
    \rowcolor[HTML]{E0E0E0} 
    \multicolumn{3}{c}{\multirow{-2}{*}{\cellcolor[HTML]{E0E0E0}{Class}}} & {$\text{mAP}_{50}$} & {IoU} & {$\text{mAP}_{50}$} & {IoU} \\ \hline
    \multicolumn{3}{c}{Person} & 0.723 & 0.707 & 0.814 & 0.789 \\ \hline
    \multicolumn{3}{c}{Car} & 0.683 & 0.655 & 0.805 & 0.766 \\ \hline
    \multicolumn{3}{c}{Truck} & 0.600 & 0.577 & 0.651 & 0.609 \\ \hline
    \multicolumn{3}{c}{Mean} & 0.667 & 0.644 & 0.756 & 0.721 \\ \hline
    \end{tabular}
    \label{yolo_res}
\end{table}

\subsection{Distance Assessor}\label{eval_assessor}
Because the final results rely heavily on the accuracy of the detected critical areas, this aspect was subjected to further analysis. To validate the distance estimation experimentally, real-world measurements were conducted and compared with algorithmically identified zones. This required direct access to a stationary tram, which was made possible through collaboration with Public Transport in Pilsen. With their support, a tram was made available at the depot. Three critical zones at distances of 60, 100, and 200 cm were defined based on recommendations from the head of ED-provoz. These zones were visually marked using orange and black tape. To closely replicate actual camera input, photographs were taken from the tram driver’s point of view. In post-processing, all other tracks in the image were masked to ensure that the segmentation focused solely on the track occupied by the reference tram (in a real world scenario, we do not focus on just the given trams track, but on the whole track bed even in the opposite direction).

\begin{figure}[!t]
    \centering
    \includegraphics[width=0.8\textwidth]{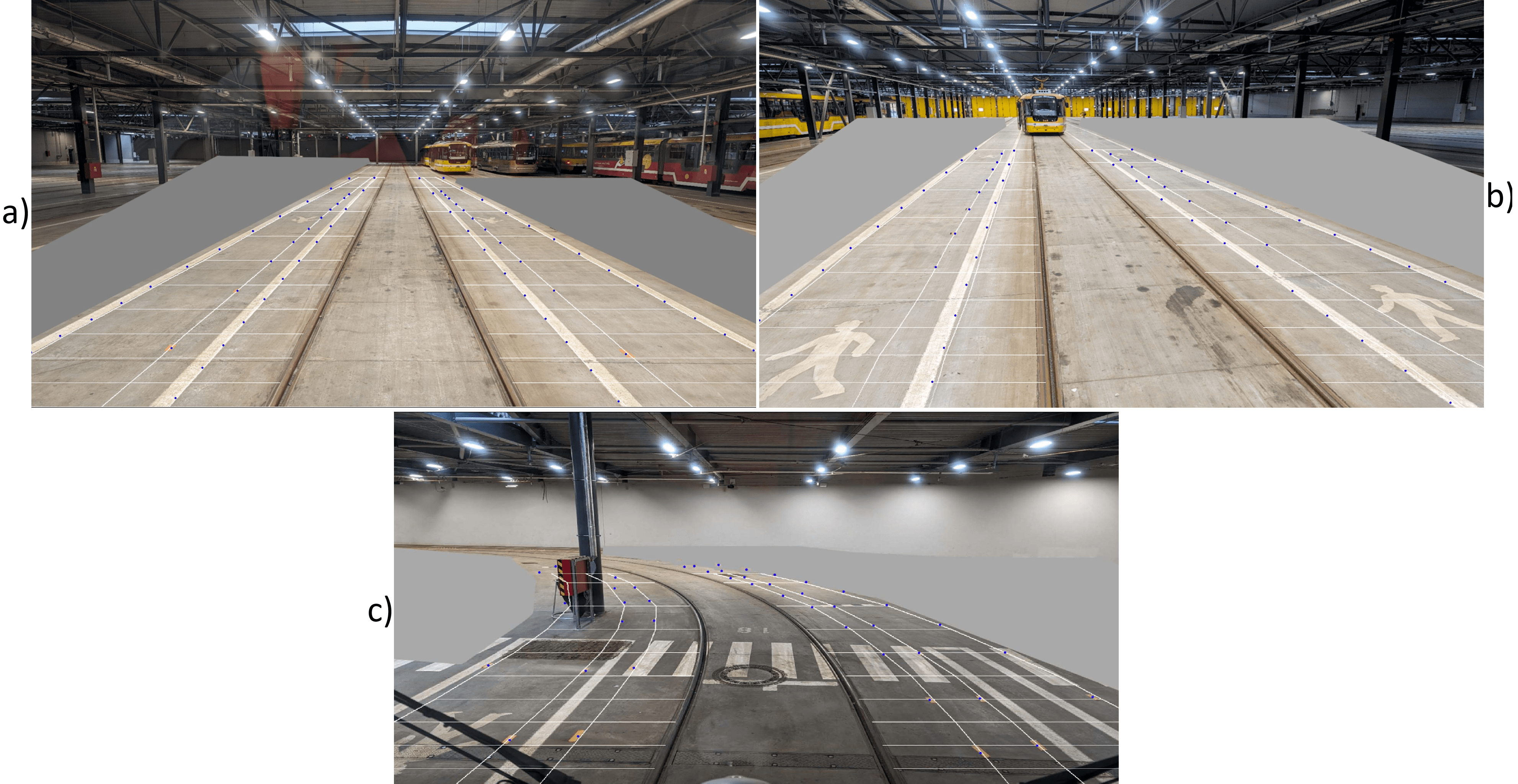}
    \caption{Examples of critical zones estimations under different simulated conditions: perpendicular view (a), oblique view (b), curved rail track (c)}
    \label{estimations_ex}
\end{figure}

Estimations shown in Figure~\ref{estimations_ex} is highly precise from a perpendicular view (a), with little variations at distant and lateral points. The second image, captured from an angle simulating a tram arriving from a turn (b), confirms the method's capability without relying strictly on straight rails, although slight deviations occur due to imperfect track segmentation. The third example, involving curved tracks (c), introduces greater complexity. Using horizontal cross-sectional measurements causes angled tracks to appear wider, slightly enlarging distance estimates. This effect is noticeable on the right side of the image, where the estimated area overlaps the actual boundary. Furthermore, narrower segmentation on the left side shifts the critical area inward, causing deviations of approximately 3-4 cm at the widest points. Despite minor accuracy reductions with sharper angles, deviations remain minimal, preserving the practical usability of the system.

\subsection{Output Visualizations}
\begin{figure}[!t]
    \centering
    \includegraphics[width=0.55\textwidth]{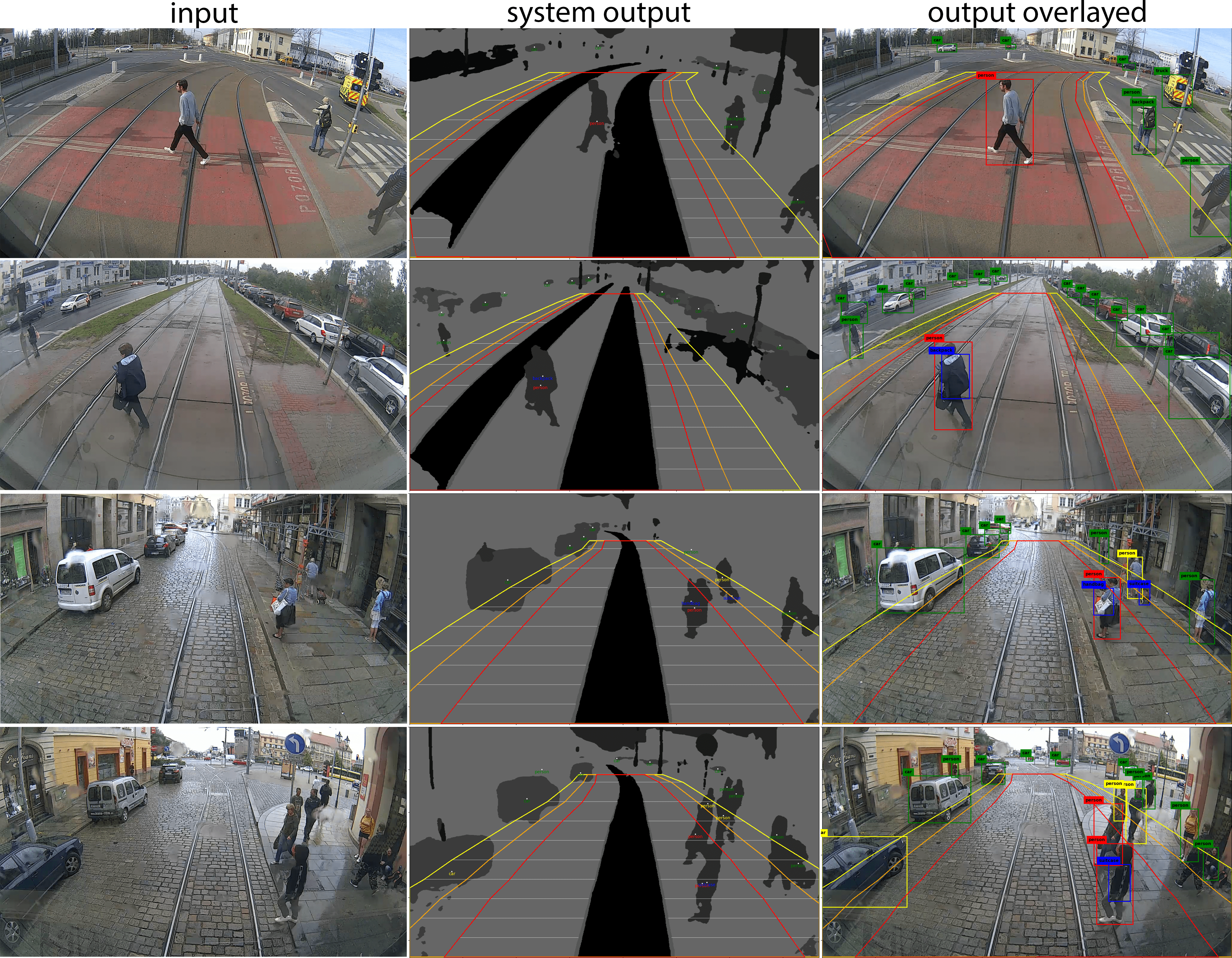}
    \includegraphics[width=0.55\textwidth]{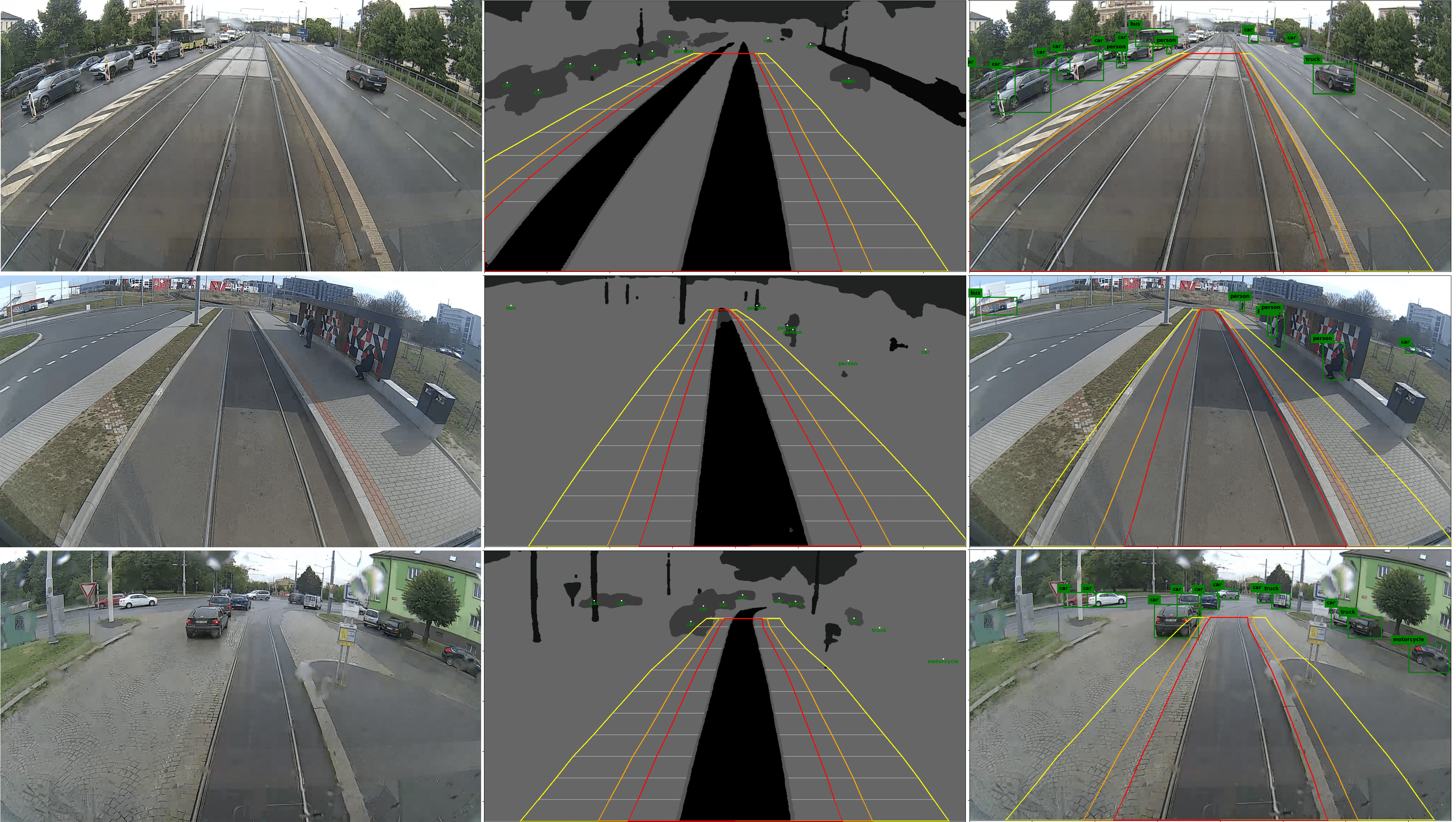}
    \caption{The final outputs from the~system, cases of~people crossing and~surrounding the~track to~show the~potential critical situations and cases of a typical traffic situations with cars surrounding tram tracks.}
    \label{outputs_figure}
\end{figure}

This study prioritized improving safety for the most frequent tram-surrounding entities — particularly pedestrians, cyclists and vehicles. Figure~\ref{outputs_figure} includes few end system outputs. 

\subsection{Additional Material}

As a final demonstration of system functionality, two visualizations are provided.
The first is a processed video showing frame-by-frame system inference
(\href{https://drive.google.com/file/d/1JHW8TfCtTTjQT848u3zUgqLuxSan3gMt/view?usp=sharing}{video}).
The second is an interactive Hugging Face Space that lets users upload images and see
risk overlays (\href{https://huggingface.co/spaces/oValach/RailSafeNet}{online demo}).

\section{Conclusion}

This paper addressed the critical issue of tram-human interaction safety by developing a comprehensive framework that integrates automatic scene segmentation, object detection, and accurate estimation of object-to-rail distances. Leveraging advanced deep learning architectures, specifically the SegFormer model, we implemented class-specific data filtration and sophisticated mask postprocessing strategies, which significantly enhanced segmentation performance. The resulting model achieved an Intersection-over-Union (IoU) score of 65.0\%, exceeding previously established benchmarks on the RailSem19 dataset. Concurrently, the object detection task was addressed by fine-tuning the YOLOv8 model on relevant classes, effectively improving its localization performance, with a mean Average Precision ($\text{mAP}_{50}$) of 75.6\% and an IoU of 72.1\%. 

A key innovation of this research is the introduction of the Distance Assessor system, designed to integrate segmentation and detection outputs effectively. This system classifies detected objects according to their criticality based on proximity to tram tracks, without the need for explicit depth data or camera parameters. Practical validation was conducted through collaborative experiments with Public Transport in Pilsen, demonstrating that the system accurately estimates distances even under challenging conditions, such as curved tracks or angled camera perspectives, with deviations limited to just a few centimeters.

Overall, this work provides robust evidence supporting the potential of combining digital image processing and artificial intelligence to significantly enhance tram safety. Further improvements will be realized through enhancements in training datasets, particularly in terms of more precise rail and rail-bed annotations. Such advancements will position the proposed framework as a practical, everyday tool for tram operators, offering the possibility to integrate this solution into some automatic breaking system etc., substantially reducing the risk and severity of accidents involving pedestrians, cyclists, and vehicles in dense urban environments.

\begin{credits}
\subsubsection{\ackname} The work has been supported by the grant of the University of West Bohemia, project No. SGS-2025-011. Computational resources were provided by the e-INFRA CZ project (ID:90254), supported by the Ministry of Education, Youth and Sports of the Czech Republic.
\subsubsection{\discintname} The authors have no competing interests to declare that are relevant to the content of this article.
\end{credits}

\bibliographystyle{splncs04}
\bibliography{database}

\end{document}